\documentclass[runningheads]{llncs}

\usepackage[T1]{fontenc}
\usepackage{graphicx}
\usepackage{amsmath}
\usepackage{amsfonts}
\usepackage{caption}
\usepackage{amssymb}
\usepackage{subcaption}
\usepackage[colorlinks=true, linkcolor=blue, urlcolor=magenta, citecolor=blue]{hyperref}
\usepackage{CJKutf8}
\usepackage{caption}
\usepackage{graphicx}
\usepackage{latexsym}
\usepackage{epstopdf}
\usepackage{float} 
\usepackage{marvosym}

\usepackage{amsmath,amsthm,amssymb,amsfonts,mathrsfs}
\usepackage{booktabs}
\usepackage{url}
\usepackage{makecell}
\usepackage{multirow}
\usepackage[section]{placeins}

\usepackage{booktabs} 
\usepackage[table]{xcolor} 

\begin{document}
\title{MEET-Sepsis: Multi-Endogenous-View Enhanced Time-Series Representation Learning for Early Sepsis Prediction}
\titlerunning{Representation Learning for Early Sepsis Prediction}

\author{Zexi Tan\inst{1}\textsuperscript{$\dagger$} \and Tao Xie\inst{1}\textsuperscript{$\dagger$} \and Binbin Sun\inst{2} \and Xiang Zhang\inst{3} \and \\ Yiqun Zhang\inst{1,4}$^\text{(\Letter)}$ \and Yiu-Ming Cheung\inst{4}$^\text{(\Letter)}$}
\authorrunning{Z. Tan \textit{et al.}}
\institute{Guangdong University of Technology, Guangzhou, China 
\and
Women and Children's Medical Center of Southern Medical University, \\Shenzhen Maternity and Child Healthcare Hospital, Shenzhen, China\\
\and
Shenzhen BH-Energy Technology Co., Ltd., Shenzhen, China\\
\and
Hong Kong Baptist University, Hong Kong SAR, China\\
\email{
yqzhang@gdut.edu.cn, ymc@comp.hkbu.edu.hk}
\footnotetext{$\dagger$ Co-first authors.\ \ \ \ $\text{\Letter}$ Corresponding author.}
}

\maketitle 

\begin{abstract}
Sepsis is a life-threatening infectious syndrome associated with high mortality in intensive care units (ICUs). Early and accurate sepsis prediction (SP) is critical for timely intervention, yet remains challenging due to subtle early manifestations and rapidly escalating mortality. While AI has improved SP efficiency, existing methods struggle to capture weak early temporal signals. This paper introduces a \textbf{M}ulti-\textbf{E}ndogenous-view \textbf{R}epresentation \textbf{E}nhancement (MERE) mechanism to construct enriched feature views, coupled with a \textbf{C}ascaded \textbf{D}ual-convolution \textbf{T}ime-series \textbf{A}ttention (CDTA) module for multi-scale temporal representation learning. The proposed \textbf{MEET-Sepsis} framework achieves competitive prediction accuracy using only 20\% of the ICU monitoring time required by SOTA methods, significantly advancing early SP. Extensive validation confirms its efficacy. Code is available at: \url{https://github.com/yueliangy/MEET-Sepsis}.
\end{abstract}

\section{Introduction}
Sepsis is a life-threatening syndrome characterized by a dysregulated immune system response to infection, leading to organ dysfunction. It can rapidly develop into septic shock within 24 hours and becomes the leading cause of death in Intensive Care Unit (ICU) patients. Therefore, Sepsis Prediction (SP) is crucial in saving lives and improving the efficiency of ICU resource allocation. However, ICU patients already have multiple diseases that can easily mask sepsis, which pose major challenges to existing SP solutions. In this context, Artificial Intelligence (AI)-empowered SP methods~\cite{kamran2024evaluation} have been developed in recent years, achieving significant advance prediction and accuracy improvements compared to manual clinical evaluation, and freeing up the workload of diagnostic experts.

\begin{figure}[!t]
\centering
\includegraphics[width=1\textwidth]{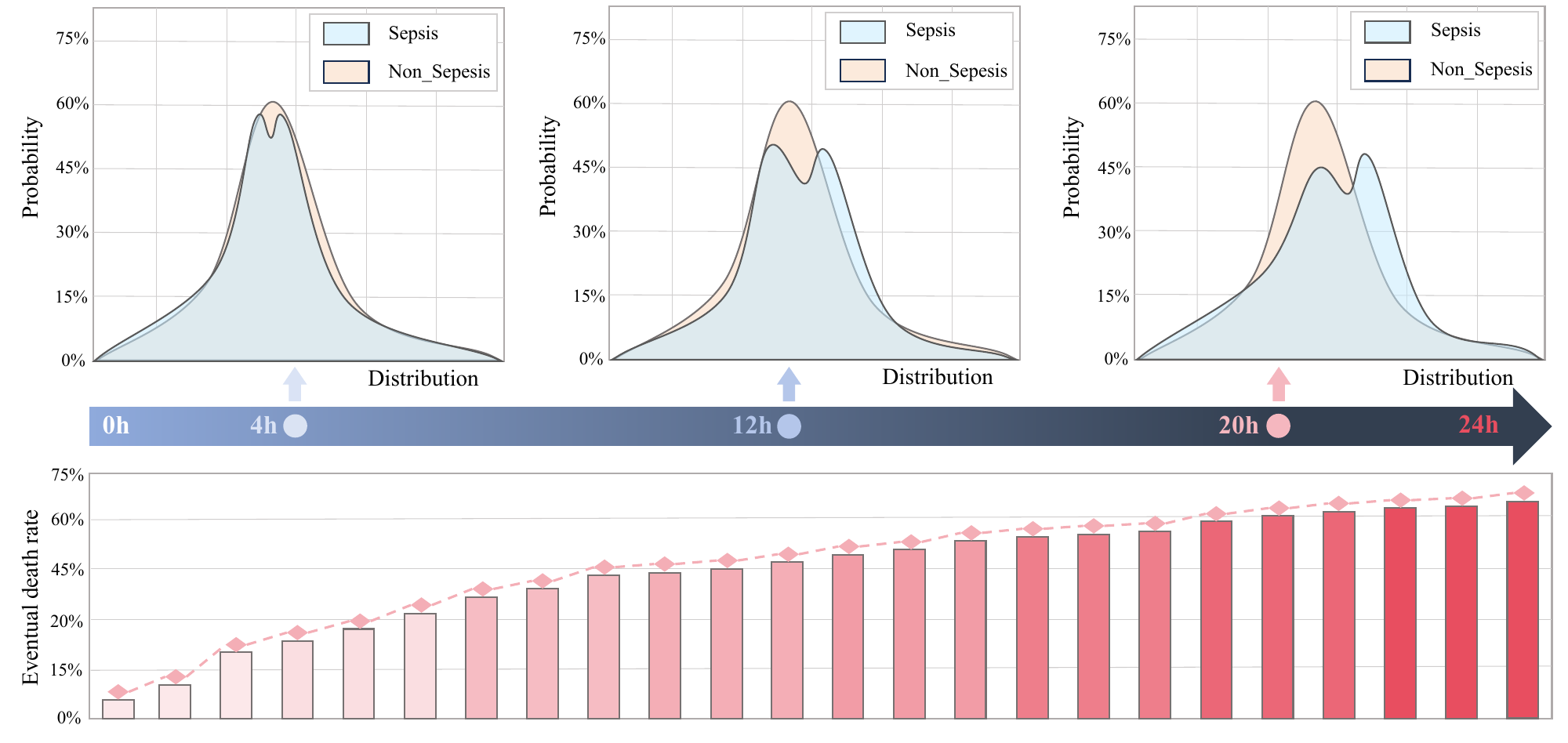}
\caption{Distribution of ICU patients in the class of ``Sepsis'' and ``Non-Sepsis'' over a 24-hour period. In the early stages (0 - 12h), the distribution difference of the two classes is subtle, while the death rate of the ``Sepsis'' class increases sharply.} \label{early}
\end{figure}

The current AI-empowered SP methods can be roughly categorized into traditional machine learning approaches~\cite{Norawit2022} and Deep Learning (DL) approaches~\cite{ZhangzhipengPRCV}. Traditional models (e.g., LR, SVM, RF, GBDT) offer interpretability, but most of them are based on static data, thus lack the ability to model complex temporal changing of sepsis. DL models~\cite{zhang2023timeseries} (e.g., LSTM and Transformer) capture temporal dependencies yet suffer from high computational cost and are short in interpretability. Hybrid approaches~\cite{ijcaiNCDE} have also been proposed, improving SP performance by fusing static and dynamic features. Nevertheless, most of the state-of-the-art solutions focus on improving the accuracy of SP on time-series data of 12 hours or longer, but lack the ability in the Early SP (ESP) task, which is actually more critical in clinical practice.

It is noteworthy that the final mortality rate of patients increases rapidly in the early stages of sepsis, as shown in Fig.~\ref{early}. However, since sepsis is a syndrome that evolves from other diseases, the early signals are extremely subtle. Moreover, the time-series records of real sepsis data are sparse, with only one or even no record per hour. These difficulties have become the accuracy bottleneck of most existing solutions using conventional time-series data representation models~\cite{GAO2024meta-analysis,ijcai2024XuHaihua}, making ESP an extremely challenging task.

This paper, therefore, proposes MEET-Sepsis, a Multi-Endogenous-view Enhanced Time-series representation learning model, to capture the early pattern of sepsis in ICU patients from weak temporal signals. It is facilitated form three aspects: 1) Feature-level convolution with nonlinear activations is adopted to construct multiple endogenous views of the input, enabling flexible exploration of inter-variable coupling; 2) Long- and short-term dependencies of time-series are captured through a cascaded dual-convolution module with kernels of different scales, and a self-attention mechanism is also introduced to dynamically weight time steps and capture the multi-scale embedding dependencies; 3) At the instance level, an ensemble prediction head is trained to partition the feature space hierarchically, improving both prediction accuracy and interpretability. Main contributions of this work are summarized into four-fold:

\begin{itemize}
\item \textbf{Practical early Sepsis prediction solution.} The proposed MEET-Sepsis method significantly improves prediction time with still competent accuracy, offering a highly practical medical time-series prediction solution.

\item \textbf{Feature enhancement.} MERE generates multiple endogenous feature views to promote sufficient non-linear feature coupling learning in CDTA to compensate for the sparsity of early temporal features.

\item \textbf{Time-series enhancement.} A dual cascaded module using multi-scale convolution kernels is adopted to extract long- and short-term dependencies, capturing the discriminative non-periodic temporal patterns of early sepsis.

\item \textbf{Reliable and safe.} 
The multi-view enhancement and the multi-scale feature extraction ensure informative representations, thereby improving the prediction reliability. Early prediction naturally leads to more time to treat, making the method safer for medical use.
\end{itemize}

\begin{figure}[!t]
\centering
\includegraphics[width=1.0\textwidth]{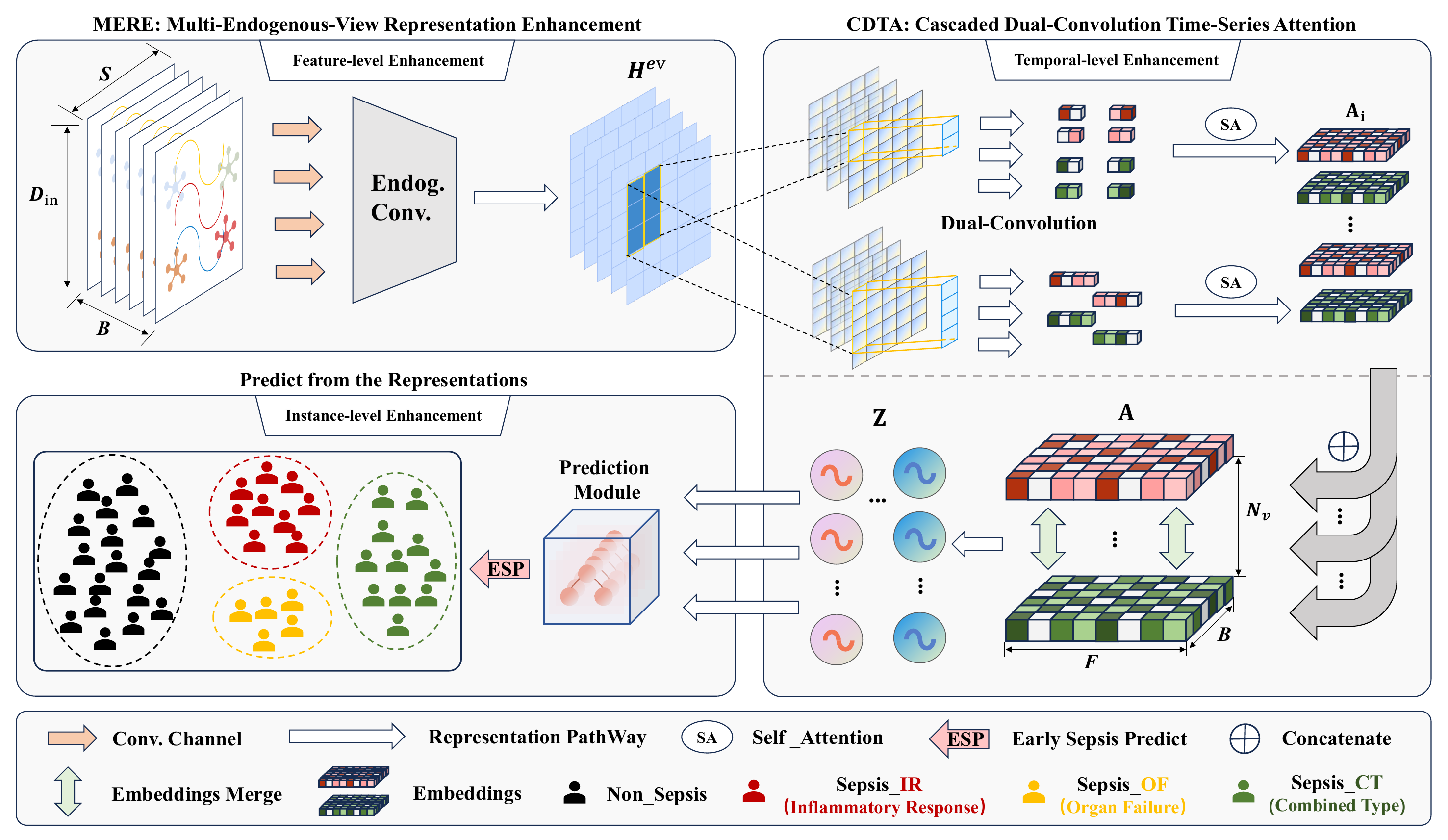}
\caption{Overview of the MEET-Sepsis framework.}
\label{famework}
\end{figure}

\section{Proposed Method}

The proposed MEET-Sepsis method comprises: 1) A MERE module that facilitates inter-feature-view coupling learning for feature enhancement, 2) A CDTA module for multi-scale temporal feature extraction, and 3) An ensemble prediction module. The overall pipeline is illustrated in Fig.~\ref{famework}.

\subsubsection{MERE: Multi-Endogenous-View Representation Enhancement.}

To effectively exploit the limited information inherent in abbreviated ICU monitoring time-series, this work employs a convolutional architecture to generate multiple endogenous feature views. These views are designed to capture intrinsic feature interdependencies within latent subspaces, thereby furnishing the subsequent CDTA module with an enriched set of complementary feature representations derived from the data's inherent structure. Formally, the temporal ICU dataset is represented as a tensor $\mathbf{X} \in \mathbb{R}^{B \times D_{\text{in}} \times S}$, where $B$, $D_{\text{in}}$, and $S$ denote the batch size, input feature dimensionality, and temporal sequence length, respectively. To augment the feature representations, a convolutional encoder is employed to project the input $\mathbf{X}$ into a set of multi-endogenous-views $\mathbf{H}^{\text{ev}}$, formulated as:

\begin{equation}
\mathbf{H}^{\text{ev}} = \sigma \left( \text{BN} \left( \sum_{g=1}^{N_v} \mathbf{W}^{(g)} \circledast \mathbf{X}^{(g)} \right) \right),
\quad \mathbf{H}^{\text{ev}} \in \mathbb{R}^{B \times N_v \times S \times V_d},
\label{eq:groupconv}
\end{equation}
where $\mathbf{X}^{(g)} \in \mathbb{R}^{B \times S}$ denotes the $g$-th grouped subspace of $\mathbf{X}$, $\mathbf{W}^{(g)} \in \mathbb{R}^{V_d \times k}$ constitutes a set of learnable one-dimensional temporal convolution kernels of length $k$, the operator $\circledast$ signifies the one-dimensional convolution operation, $\text{BN}(\cdot)$ represents batch normalization, and $\sigma(\cdot)$ corresponds to the GELU activation function~\cite{GELU2018}. This mechanism effectively transforms the original features into $N_v$ distinct views, each comprising $V_d$-dimensional embeddings. This process facilitates enhanced representation learning for each original feature and promotes comprehensive feature interaction learning within the CDTA module.

\subsubsection{CDTA: Cascaded Dual-Convolution Time-Series Attention.} To capture long- and short-term temporal features, each view $H^{\text{ev}}[:, i, :, :], i \in \{1, \dots, N_v\}$ is input into convolution layers \( \mathcal{F}^{(l)} \) and \( \mathcal{F}^{(s)} \) with different sized kernels:
{\small
\begin{equation}
\mathbf{C}^{(l)}_i = \text{MP}(\mathcal{F}^{(l)} (H^{\text{ev}}[:, i, :, :]))
\in \mathbb{R}^{B \times F_l \times \left\lfloor \frac{S}{p} \right\rfloor}\ \ \text{and}\ \ 
\mathbf{C}^{(s)}_i = \text{AP}(\mathcal{F}^{(s)}(\mathbf{C}^{(l)}_i)) \in \mathbb{R}^{B \times F_s},
\label{eq:temporal_encoding}
\end{equation}}

\noindent where \( \mathcal{F}^{(l)}: \mathbb{R}^{B \times S \times V_d} \rightarrow \mathbb{R}^{B \times F_l \times S} \) captures long-term dependencies with kernel size \( k_1 \), MP is a max-pooling operator with stride \( p \), 
\( \mathcal{F}^{(s)} \) adopts a similar structure with kernel size \( k_2 \) ($k_2<k_1$) to extract short-term patterns, and 
AP denotes adaptive average pooling for global temporal summarization. A self-attention layer \( \mathrm{SA} \) is also adopted to enhance key patterns by:
\begin{equation}
\mathbf{A}_i = \mathrm{SA}(\mathbf{C}^{(s)}_i W_Q^h, \mathbf{C}^{(s)}_i W_K^h, \mathbf{C}^{(s)}_i W_V^h),
\end{equation}
where the input is projected into key, query, and value spaces of equal dimension, and $h$ is the number of heads. The outputs from all \( N_v \) views are concatenated to form a unified high-dimensional representation $\mathbf{A} = [\mathbf{A}_1, \mathbf{A}_2, \dots, \mathbf{A}_{N_v}] 
\in \mathbb{R}^{B \times (N_v \cdot F_s)}$. Finally, a linear projection layer generates the final fused representation for sepsis prediction:
\begin{equation}
\mathbf{Z} = \mathbf{A} \mathbf{W_p} + \mathbf{b}_p 
\in \mathbb{R}^{B \times D_{\text{proj}}},
\label{eq:linear_proj}
\end{equation}
where \( \mathbf{W_p} \in \mathbb{R}^{(N_v \cdot F_s)\times D_{\text{proj}}} \) and \( \mathbf{b}_p \in \mathbb{R}^{D_{\text{proj}}} \) are the learnable parameters.

\subsubsection{MEET-Sepsis: Model Training.}

An alternating optimization strategy is adopted to train the whole model. The loss function is decomposed into two sub-objectives: 1) Reconstruction MSE loss $\mathcal{L}_{MSE}$ with $\ell_2$ regularization term $\mathcal{L}_{reg}$ to preserve consistency between the learned representation with the original feature space, and 2) prediction loss $\mathcal{L}_{pred}$ to adapt the prediction module to the labels, which can be written with the trade-off parameters $\alpha$ and $\beta$ as:
\begin{equation}
\mathcal{L} = \mathcal{L}_{MSE} + \alpha \mathcal{L}_{reg} + \beta \mathcal{L}_{pred}.
\label{eq:loss}
\end{equation}
The alternative training is implemented using Adam optimizer, and the obtained representation $\mathbf{Z}$ is utilized for early spesis prediction.

\section{Experiments}
\subsubsection{Experimental Settings.}
The effectiveness of MEET-Sepsis is evaluated by comparison with baseline methods (\textit{XGBoost}~\cite{XGBoost2016}, \textit{RandomForest}~\cite{Pal2005}) and state-of-the-art approaches (\textit{TimeMixer}~\cite{wang2023timemixer}, \textit{MPTSFNet}~\cite{mu2025mptsnet}, \textit{OneStepKNN}~\cite{Zhang2023}), as well as through ablation studies to validate key components. Experiments use two ICU datasets from PhysioNet/CinC Challenge 2019~\cite{reyna2020early}: qSOFA (10,856 samples, 4-class) and SOFA (12,259 samples, 3-class), labeled per clinical criteria. Each record is divided into 22 hourly segments (2--23 hours after ICU admission) to model sepsis progression and facilitate evaluation under increasing observation windows. Performance is measured via Accuracy and F1-Score metrics. To ensure statistical reliability, all experiments were conducted with five independent runs, and the results are reported as mean ± standard deviation. Key hyperparameters for MEET-Sepsis include 35 endogenous views, and convolutional kernels of sizes 5 and 3. The other parameters for all the compared methods are set following the recommendations in the source literature.

\begin{figure}[!t]
\centering
\includegraphics[width=\textwidth]{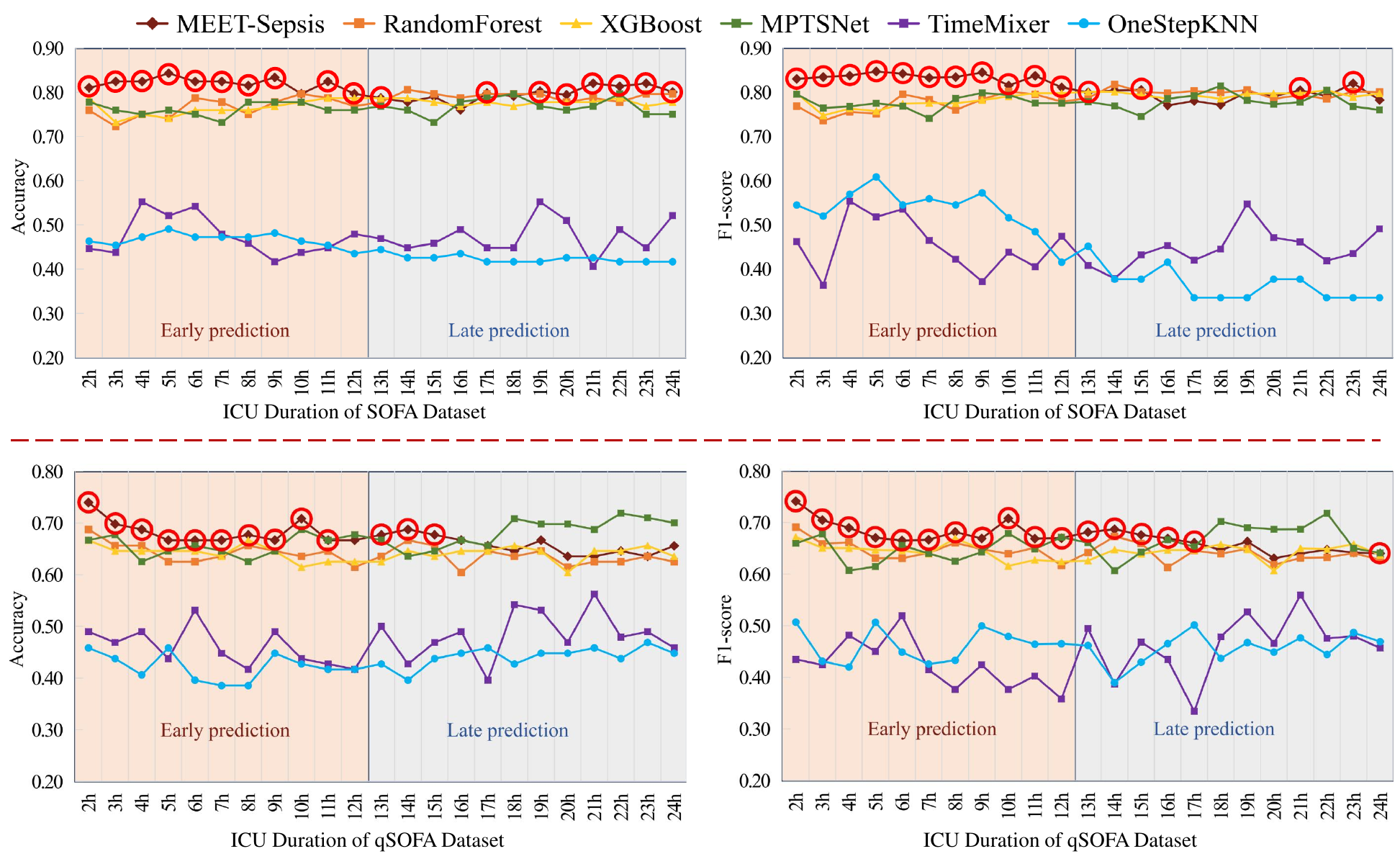}
\caption{MEET-Sepsis performance on SOFA (upper) and qSOFA (lower) datasets. The time slots when MEET-Sepsis ranked 1st are marked with \textbf{\textcolor{red}{red circles}}.
} \label{SOFA_and_qSOFA}
\end{figure}

\subsubsection{Prediction Performance Evaluation.}
As illustrated in Fig.~\ref{SOFA_and_qSOFA}, MEET-Sepsis demonstrates strong overall performance. Notably, it achieves top rankings across the majority of its results, particularly during the early prediction stages, as visually denoted by red circles. Key observations include: (1) On the SOFA dataset at the 5th hour, the proposed method exhibits a substantial advantage, attaining peak global performance within approximately 20\% of the typical monitoring duration; (2) Following the established early sepsis prediction (ESP) criterion (fewer than 12 hours), MEET-Sepsis dominates ESP tasks, ranking first in 83 out of 88 comparative scenarios (2 datasets × 4 metrics × 11 early time slots); (3) State-of-the-art baselines such as OneStepKNN and TimeMixer underperform significantly, as early weak signals prove inadequate for driving longer-term temporal models; (4) Even in later prediction intervals, although primarily designed for ESP, MEET-Sepsis maintains highly competitive performance, consistently ranking among the top three, highlighting its robustness and generalizability across various sepsis progression stages; (5) MEET-Sepesis exhibits consistent performance stability across temporal segments and evaluation metrics, whereas strong competitors (e.g., MPTSNet and XGBoost) display considerable temporal performance fluctuations.

\subsubsection{Ablation Study.}
To evaluate the contribution of the core MERE and CDTA modules, this paper conducts ablation experiments on the SOFA dataset at the 5-hour time slot, where MEET-Sepsis demonstrates particularly strong performance as shown in Fig.~\ref{SOFA_and_qSOFA}. Three ablated variants are compared: without MERE (w/o MERE), without CDTA (w/o CDTA), and without both modules (w/o both). Results in Table~\ref{ablation} reveal that removing either module leads to consistent performance deterioration across both accuracy and F1-score. Furthermore, the complete removal of both modules reduces the model to an XGBoost-based baseline, which underperforms all partial variants. These results affirm that both modules are indispensable and operate synergistically to enhance predictive capability. The progressive performance decline across ablation settings underscores the importance of their integrated design in capturing complementary feature interactions and temporal dynamics.

\begin{table*}[!t]
    \centering
    \caption{Ablation study on the SOFA dataset at time slot 5h. The results marked in \textbf{\textcolor{orange}{Orange}} color indicate the best performance in terms of each index.}
    \label{ablation}
    \resizebox{1\textwidth}{!}{
        \renewcommand{\arraystretch}{1.3}
        \setlength{\tabcolsep}{10pt}
        \begin{tabular}{l|c|ccc}
            \toprule
            & MEET-Sepsis & w/o MERE & w/o CDTA & w/o both \\
            \midrule
            Accuracy & \cellcolor{orange!40}0.8426±0.0277 & 0.7500±0.0139 & 0.7593±0.0203 & 0.7407±0.0093 \\
            F1-score & \cellcolor{orange!40}0.8468±0.0295 & 0.7548±0.0141 & 0.7680±0.0219 & 0.7572±0.0106 \\
            \bottomrule
        \end{tabular}
    }
\end{table*}

\subsubsection{Discussions.} Due to space constraints, several merits of MEET-Sepsis have been empirically observed outside the paper and are summarized as follows: 1) \textbf{Scalability:} The model does not introduce much extra computational overhead compared to mainstream alternatives, owing to its efficient architecture without extremely deep layers. 2) \textbf{Stability:} The incorporation of multi-view and multi-scale temporal feature extraction contributes to robust and consistent performance. 3) \textbf{Extensibility:} The framework shows strong potential for extension to federated learning scenarios~\cite{rieke2020future,Zhangyunfan} and multi-modal heterogeneous data integration~\cite{Chen2024QGRL,Zhao2024DM}, facilitating future applications in privacy-aware collaborative learning and complex clinical environments.

\section{Concluding Remarks}
This paper proposes MEET-Sepsis, a novel early sepsis prediction approach that enhances representation learning from weak temporal signals to accurately predict sepsis. By integrating the MERE module for feature enhancement and the CDTA module for time-series enhancement, the prediction head can be effectively trained to distinguish between sepsis and non-sepsis cases, significantly advancing prediction time without sacrificing accuracy. Experiments confirm the promising performance. Nevertheless, MEET-Sepsis has certain limitations. The fixed truncation of time-series may not adapt to the variable progression rates of sepsis across patients, potentially losing critical early information. Moreover, the generated endogenous views may not fully align with the true semantic distribution of the data, which could impose certain constraints on the model's generalization capability. Future work will explore adaptive time windows and improved augmentation strategies to mitigate these issues.

\section{Acknowledgement}
This work was supported in part by the National Natural Science Foundation of China (NSFC) under Grant 62476063, the NSFC/Research Grants Council (RGC) Joint Research Scheme under Grant N\_HKBU214/21, the Natural Science Foundation of Guangdong Province under Grant 2025A1515011293, the General Research Fund of RGC under Grants 12202622 and 12201323, the RGC Senior Research Fellow Scheme under Grant SRFS23242S02, the General Projects of Shenzhen Science and Technology Program under Grant JCYJ202408
13115124032, and the Shenzhen Maternity and Child Healthcare Hospital under Grant FYA2022018.

\bibliographystyle{splncs04}
\bibliography{mybib}

\end{document}